\pdfoutput=1

\documentclass[11pt]{article}

\usepackage[final]{acl}

\usepackage{times}
\usepackage{latexsym}

\usepackage[T1]{fontenc}

\usepackage[utf8]{inputenc}

\usepackage{microtype}

\usepackage{inconsolata}

\usepackage{graphicx}

\author{
Zixin Tang\textsuperscript{1}~~~Chieh-Yang Huang\textsuperscript{2}~~~Tsung-Che Li\textsuperscript{3}~~~Ho Yin Sam Ng\textsuperscript{1}\\
\textbf{Hen-Hsen Huang\textsuperscript{3}~~~Ting-Hao `Kenneth' Huang\textsuperscript{1}}\\
\textsuperscript{1}College of Information Sciences and Technology, The Pennsylvania State University\\
\textsuperscript{2}MetaMetrics Inc.~~~\textsuperscript{3}Institute of Information Science, Academia Sinica\\
\textsuperscript{1}\texttt{\{zxtang,sam.ng,txh710\}@psu.edu}~~~\textsuperscript{2}\texttt{cyhuang@lexile.com} \\
\textsuperscript{3}\texttt{\{george,hhhuang\}@iis.sinica.edu.tw}
}

\usepackage{xspace}
\usepackage{comment}
\usepackage{multirow}
\usepackage{booktabs}
\usepackage{multirow}

\usepackage{amsmath}
\usepackage{cleveref}
\usepackage{booktabs}
\usepackage{multirow}

\usepackage{CJKutf8}
\usepackage{siunitx}
\usepackage{listings}
\usepackage{tabularx}
\usepackage{enumitem}

\usepackage{soul}

\PassOptionsToPackage{table,xcdraw}{xcolor}
\usepackage{colortbl}

\definecolor{lightred}{RGB}{255,200,200}
\definecolor{lightblue}{RGB}{200,200,255}
\definecolor{lightgreen}{RGB}{200,255,200}
\definecolor{lightyellow}{RGB}{255,255,200}
\definecolor{eclipseStrings}{RGB}{42,0.0,255}
\definecolor{eclipseKeywords}{RGB}{127,0,85}
\colorlet{numb}{magenta!60!black}

\title{Using Contextually Aligned Online Reviews to\\Measure LLMs' Performance Disparities Across Language Varieties}

\newcommand{\kenneth}[1]{}
\newcommand{\zixin}[1]{}
\newcommand{\sam}[1]{}
\newcommand{\cy}[1]{}

\newcommand{\eg}{{\it e.g.}\xspace}
\newcommand{\ie}{{\it i.e.}\xspace}

\newcommand{\twChinese}{{T}aiwan {M}andarin\xspace}
\newcommand{\cnChinese}{{M}ainland {M}andarin\xspace}

\begin{document}

\maketitle



\begin{abstract}
A language can have different varieties. 
These varieties can affect the performance of natural language processing (NLP) models, including large language models (LLMs), which are often trained on data from widely spoken varieties.
This paper introduces a novel and cost-effective approach to benchmark model performance across language varieties.
We argue that international online review platforms, such as Booking.com, can serve as effective data sources for constructing datasets that capture \textbf{comments in different language varieties from similar real-world scenarios}, like reviews for the same hotel with the same rating using the same language (\eg, Mandarin Chinese) but different language varieties (\eg, \twChinese, \cnChinese).  
To prove this concept, we constructed a \textbf{contextually aligned} dataset comprising reviews in \twChinese and \cnChinese and tested six LLMs in a sentiment analysis task. 
Our results show that LLMs consistently underperform in \twChinese. 
\end{abstract}



\section{Introduction}

A language can have different varieties. 
Of the world's 7,000 languages, sixty (60) million people speak British English, 23 million speak \twChinese, and 10 million speak European Portuguese, compared to 330 million, 900 million, and 200 million who speak American English, \cnChinese, and Brazilian Portuguese, respectively. 
These varieties differ enough in accent, vocabulary, or syntax for native speakers to distinguish them. 
NLP technologies, including LLMs, are known to perform better in English varieties that are more widely represented in the internet data they are trained on, particularly Mainstream American English (MAE), compared to less represented varieties like African American English (AAE)~\cite{valuebench,multivaluebench}.
Specifically, LLMs more accurately predict sentiment scores in MAE~\cite{valuebench}, 
generate higher-quality texts in MAE~\cite{valuebench}, and 
hold better conversations in MAE~\cite{multivaluebench}.
These comparisons were made possible by intensive, targeted efforts specific to each language variety, such as 
``translating'' data instances from a standard variety (\eg, MAE) to less widely represented varieties (\eg, AAE), followed by validation from native speakers~\cite{valuebench,multivaluebench}.
What is not known is whether these performance gaps and biases extend to a broader range of languages and their numerous varieties, 
such as \cnChinese versus \twChinese.
Building effective benchmarking datasets for evaluating model performance across language varieties is expensive---creating ``fair'' comparisons between varieties often needs native speakers and language experts.

\begin{figure}[t]
    \centering
    \includegraphics[width=.9\columnwidth]{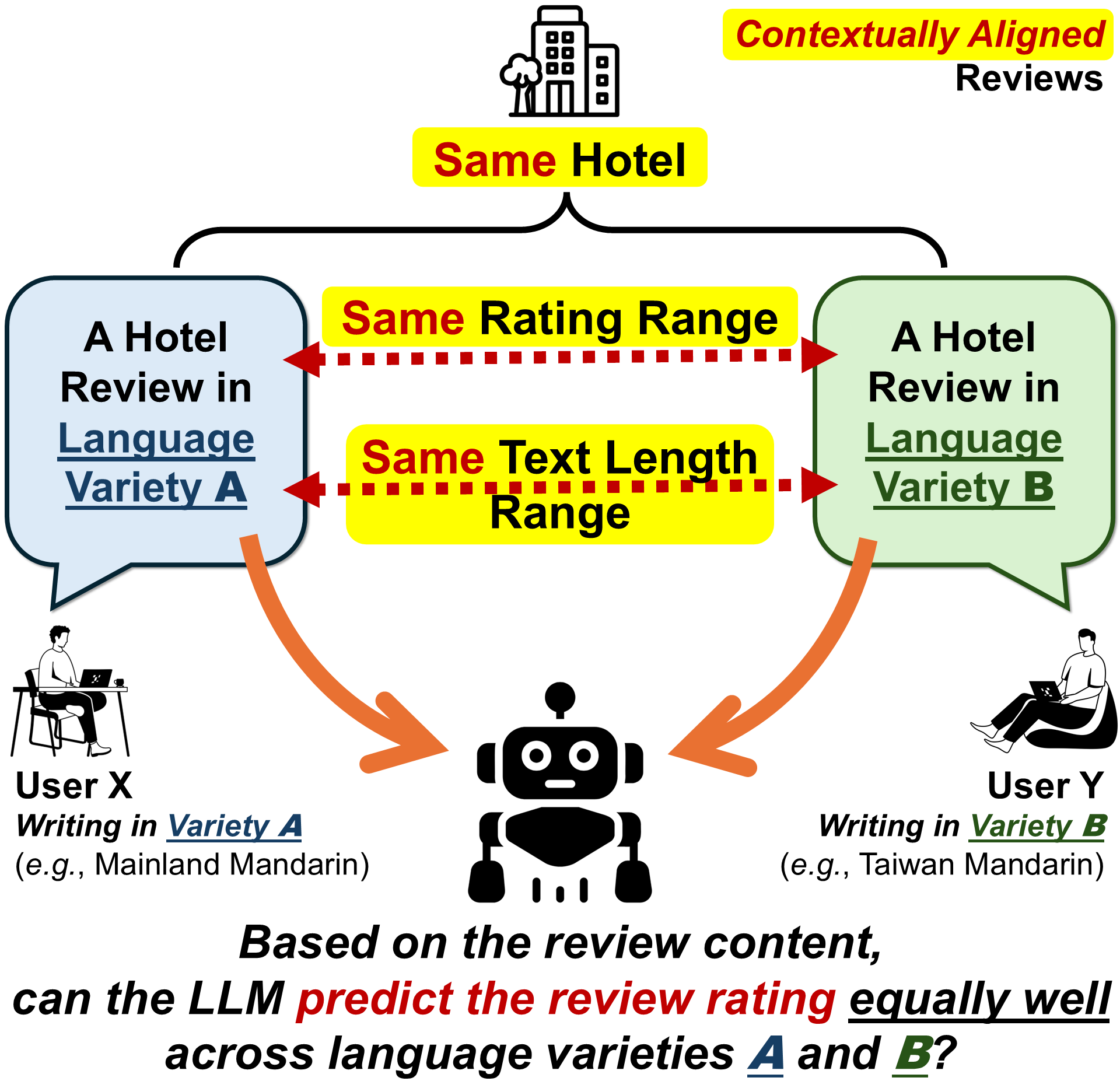}
    \vspace{-.5pc}
    \caption{Online review platforms can be data sources to build datasets that capture comments in different language varieties from similar real-world scenarios. These \textit{contextually aligned} datasets can then be used to benchmark LLMs' performance across language varieties.}
    \vspace{-1.5pc}
    \label{fig:overview}
\end{figure}

Using Mandarin Chinese as an example, we propose an approach that uses large-scale user-generated reviews to construct benchmarking datasets across varieties of a given language. 
We argue that the international online review platforms with millions of users, like Booking.com, when properly curated, can serve as effective data sources for constructing datasets that capture \textbf{comments in different language varieties from similar real-world scenarios}, like comments for the same hotel with the same rating using the same language (\eg, Mandarin Chinese) but different language varieties (\eg, \twChinese, \cnChinese). 
These datasets, being \textbf{contextually aligned}, can then be used to benchmark LLMs' performance across language varieties for tasks like sentiment analysis and text generation (Figure~\ref{fig:overview}). 
Once a low-cost and generalizable approach becomes available, researchers can then compare model performance across a wide range of language varieties, enabling reliable benchmarking of progress in addressing performance gaps and moving toward an LLM that performs equally well across all language varieties. 






\section{Related Work}

Beyond machine translation~\cite{kantharuban2023quantifying},
researchers tried to benchmark NLP models across language varieties~\cite{Zampieri_Nakov_Scherrer_2020,joshi2024natural,blodgett2020biasreview,hovy-johannsen-2016-exploring,vardial2019report}, but the focus on identifying gaps between these varieties varies widely.
Some prior work focused solely on a single less-representative variety, such as \twChinese~\cite{tamtmmlu+,chen2024measuring}, without measuring performance gaps across multiple varieties.
Other studies that measured these gaps employed different levels of granularity.
The most common approach, \textbf{task-level comparison}, benchmarks the same NLP task across language varieties~\cite{dialectbench}, such as sentiment analysis, but datasets often differ in source or genre across varieties, making the reported performance numbers not directly comparable. 
For instance, sentiment analysis datasets for \cnChinese and \twChinese often used different sources~\cite{seki2007overview}. 
A more refined approach, \textbf{scenario-level comparison}, evaluates performance within the same dataset or scenario, 
such as essay grading~\cite{liang2023gptesl} or speech rating~\cite{kwako2023bertbias},
across data partitions of different language varieties~\cite{lwowski2022measuring,AAE2017racial}. 
While this method eliminates biases caused by differing data sources, it cannot fully address biases introduced during dataset construction.
The most rigorous method, \textbf{instance-level comparison}, involves constructing parallel datasets with an item-by-item alignment between varieties~\cite{valuebench,multivaluebench,groenwold2020AAENLP,macucocorpus}, where each instance is converted between language varieties. 
However, creating such comparisons is very costly, requiring native speakers and language experts to ensure accuracy. 
Our approach achieves instance-level comparability with lower costs.





\begin{table*}[t]
\centering
\footnotesize
\begin{tabular}{@{}ccccr@{\kern-4mm}lccr@{\kern-4mm}lccr@{\kern-4mm}l@{}}
\toprule
 &  & \multicolumn{12}{c}{\textbf{Accuracy (Acc)$\uparrow$}} \\ \cmidrule{3-14}
 &  & \multicolumn{4}{c}{\textbf{structured}} & \multicolumn{4}{c}{\textbf{plain}} & \multicolumn{4}{c}{\textbf{shuffled}} \\ \cmidrule(lr){3-6} \cmidrule(lr){7-10} \cmidrule(lr){11-14}
\multirow{-4}{*}{\textbf{\begin{tabular}[c]{@{}c@{}}Text\\ Length\\ (\#Character)\end{tabular}}} & \multirow{-4}{*}{\textbf{Model}} & \textbf{tw} & \textbf{cn} & \multicolumn{2}{c}{\textbf{\begin{tabular}[c]{@{}c@{}}$\Delta$Acc\\ (cn-tw)\end{tabular}}} & \textbf{tw} & \textbf{cn} & \multicolumn{2}{c}{\textbf{\begin{tabular}[c]{@{}c@{}}$\Delta$Acc\\ (cn-tw)\end{tabular}}} & \textbf{tw} & \textbf{cn} & \multicolumn{2}{c}{\textbf{\begin{tabular}[c]{@{}c@{}}$\Delta$Acc\\ (cn-tw)\end{tabular}}} \\ \midrule
 & \textbf{GPT-4o} & 26.52 & 27.43 & \cellcolor[HTML]{FBE8E6}0.91\hspace{4mm} & \multicolumn{1}{l}{} & 19.16 & 20.78 & \cellcolor[HTML]{F7D5D2}1.62\hspace{4mm} & *** & 18.57 & 20.16 & \cellcolor[HTML]{F8D6D3}1.60\hspace{4mm} & *** \\
 & \textbf{Llama3 8b} & 27.40 & 26.39 & \cellcolor[HTML]{D9F1F3}-1.01\hspace{4mm} & \multicolumn{1}{l}{} & 19.21 & 19.08 & \cellcolor[HTML]{FAFDFD}-0.13\hspace{4mm} & \multicolumn{1}{l}{} & 17.43 & 17.71 & \cellcolor[HTML]{FEF8F8}0.28\hspace{4mm} & \multicolumn{1}{l}{} \\
 & \textbf{Llama3 70b} & 35.43 & 35.00 & \cellcolor[HTML]{EFF9FA}-0.43\hspace{4mm} & \multicolumn{1}{l}{} & 28.21 & 29.60 & \cellcolor[HTML]{F9DBD9}1.39\hspace{4mm} & ** & 27.54 & 29.51 & \cellcolor[HTML]{F6CCC8}1.97\hspace{4mm} & *** \\
 & \textbf{Llama3 405b} & 37.96 & 40.51 & \cellcolor[HTML]{F3BDB8}2.55\hspace{4mm} & *** & 27.42 & 30.12 & \cellcolor[HTML]{F2B9B4}2.70\hspace{4mm} & *** & 27.59 & 30.17 & \cellcolor[HTML]{F3BCB7}2.58\hspace{4mm} & *** \\
 & \textbf{Gemma2 9b} & 15.69 & 14.45 & \cellcolor[HTML]{D1EEF0}-1.24\hspace{4mm} & ** & 17.01 & 17.26 & \cellcolor[HTML]{FEF9F9}0.25\hspace{4mm} & \multicolumn{1}{l}{} & 15.81 & 16.35 & \cellcolor[HTML]{FDF1F0}0.54\hspace{4mm} & \multicolumn{1}{l}{} \\
\multirow{-6}{*}{\textbf{\begin{tabular}[c]{@{}c@{}}Short\\ (1-49)\end{tabular}}} & \textbf{Gemma2 27b} & 15.34 & 14.27 & \cellcolor[HTML]{D7F0F2}-1.07\hspace{4mm} & ** & 13.94 & 14.03 & \cellcolor[HTML]{FFFDFD}0.09\hspace{4mm} & \multicolumn{1}{l}{} & 13.91 & 14.29 & \cellcolor[HTML]{FEF6F5}0.37\hspace{4mm} & \multicolumn{1}{l}{} \\ \midrule
 & \textbf{GPT-4o} & 35.59 & 38.39 & \cellcolor[HTML]{F2B7B2}2.79\hspace{4mm} & *** & 28.15 & 33.16 & \cellcolor[HTML]{E67C73}5.01\hspace{4mm} & *** & 26.73 & 31.36 & \cellcolor[HTML]{E8867E}4.64\hspace{4mm} & *** \\
 & \textbf{Llama3 8b} & 25.31 & 27.01 & \cellcolor[HTML]{F7D3D0}1.70\hspace{4mm} & * & 19.53 & 21.24 & \cellcolor[HTML]{F7D3D0}1.71\hspace{4mm} & ** & 18.92 & 21.11 & \cellcolor[HTML]{F5C6C2}2.19\hspace{4mm} & *** \\
 & \textbf{Llama3 70b} & 34.66 & 38.24 & \cellcolor[HTML]{EEA29B}3.59\hspace{4mm} & *** & 35.02 & 37.45 & \cellcolor[HTML]{F3C0BC}2.43\hspace{4mm} & ** & 33.66 & 36.43 & \cellcolor[HTML]{F2B7B2}2.77\hspace{4mm} & *** \\
 & \textbf{Llama3 405b} & 37.20 & 40.52 & \cellcolor[HTML]{EFA9A3}3.31\hspace{4mm} & *** & 36.09 & 38.00 & \cellcolor[HTML]{F6CECA}1.91\hspace{4mm} & * & 34.38 & 36.60 & \cellcolor[HTML]{F4C5C1}2.22\hspace{4mm} & ** \\
 & \textbf{Gemma2 9b} & 14.84 & 15.66 & \cellcolor[HTML]{FBEAE9}0.82\hspace{4mm} & \multicolumn{1}{l}{} & 18.22 & 20.00 & \cellcolor[HTML]{F7D1CE}1.78\hspace{4mm} & ** & 16.59 & 17.98 & \cellcolor[HTML]{F9DBD9}1.38\hspace{4mm} & * \\
\multirow{-6}{*}{\textbf{\begin{tabular}[c]{@{}c@{}}Long\\ (50+)\end{tabular}}} & \textbf{Gemma2 27b} & 13.44 & 14.52 & \cellcolor[HTML]{FAE3E1}1.08\hspace{4mm} & \multicolumn{1}{l}{} & 15.48 & 16.99 & \cellcolor[HTML]{F8D8D5}1.51\hspace{4mm} & * & 15.16 & 17.16 & \cellcolor[HTML]{F6CBC8}2.00\hspace{4mm} & *** \\ \midrule
 & \textbf{GPT-4o} & 29.61 & 31.16 & \cellcolor[HTML]{F8D7D4}1.55\hspace{4mm} & *** & 22.22 & 24.99 & \cellcolor[HTML]{F2B7B2}2.78\hspace{4mm} & *** & 21.35 & 23.98 & \cellcolor[HTML]{F2BBB6}2.63\hspace{4mm} & *** \\
 & \textbf{Llama3 8b} & 26.69 & 26.61 & \cellcolor[HTML]{FCFDFE}-0.08\hspace{4mm} & \multicolumn{1}{l}{} & 19.32 & 19.82 & \cellcolor[HTML]{FDF2F2}0.50\hspace{4mm} & \multicolumn{1}{l}{} & 17.94 & 18.88 & \cellcolor[HTML]{FBE7E5}0.94\hspace{4mm} & * \\
 & \textbf{Llama3 70b} & 35.16 & 36.10 & \cellcolor[HTML]{FBE7E5}0.94\hspace{4mm} & * & 30.53 & 32.27 & \cellcolor[HTML]{F7D2CF}1.75\hspace{4mm} & *** & 29.62 & 31.87 & \cellcolor[HTML]{F4C5C1}2.24\hspace{4mm} & *** \\
 & \textbf{Llama3 405b} & 37.70 & 40.51 & \cellcolor[HTML]{F1B6B1}2.81\hspace{4mm} & *** & 30.39 & 32.82 & \cellcolor[HTML]{F3C0BC}2.43\hspace{4mm} & *** & 29.92 & 32.38 & \cellcolor[HTML]{F3BFBB}2.46\hspace{4mm} & *** \\
 & \textbf{Gemma2 9b} & 15.40 & 14.86 & \cellcolor[HTML]{EBF7F8}-0.54\hspace{4mm} & \multicolumn{1}{l}{} & 17.42 & 18.19 & \cellcolor[HTML]{FCEBEA}0.77\hspace{4mm} & * & 16.07 & 16.90 & \cellcolor[HTML]{FBEAE8}0.83\hspace{4mm} & * \\
\multirow{-6}{*}{\textbf{Overall}} & \textbf{Gemma2 27b} & 14.69 & 14.35 & \cellcolor[HTML]{F2FAFB}-0.34\hspace{4mm} & \multicolumn{1}{l}{} & 14.47 & 15.04 & \cellcolor[HTML]{FDF1F0}0.57\hspace{4mm} & \multicolumn{1}{l}{} & 14.34 & 15.27 & \cellcolor[HTML]{FBE7E6}0.93\hspace{4mm} & ** \\ \bottomrule
\end{tabular}
\vspace{-.5pc}
\caption{Accuracy (Acc~$\uparrow$) by length for GPT-4o, Llama3 (8b, 70b, 405b), and Gemma2 (9b, 27b) models. Red (green) indicates better (worse) performance in CN, with darker shades representing larger gaps. 
(Statistical group differences are indicated as ${^{*}}$  (p<.05), ${^{**}}$ (p<.01), and ${^{***}}$ (p<.001) regarding the model performance.)}
\vspace{-1pc}
\label{tab:new-acc-results}
\end{table*}

\begin{table*}[t]
\centering
\footnotesize
\begin{tabular}{c@{\kern2mm}c@{\kern2mm}ccr@{\kern-4mm}lccr@{\kern-4mm}lccr@{\kern-4mm}l}
\toprule
 &  & \multicolumn{12}{c}{\textbf{Mean Squared Error (MSE)~$\downarrow$}} \\ \cmidrule(l){3-14} 
 &  & \multicolumn{4}{c}{\textbf{structured}} & \multicolumn{4}{c}{\textbf{plain}} & \multicolumn{4}{c}{\textbf{shuffled}} \\ \cmidrule(lr){3-6} \cmidrule(lr){7-10} \cmidrule(lr){11-14}
\multirow{-4}{*}{\textbf{\begin{tabular}[c]{@{}c@{}}Text\\ Length\\ (\#Character)\end{tabular}}} & \multirow{-4}{*}{\textbf{Model}} & \textbf{tw} & \textbf{cn} & \textbf{\begin{tabular}[c]{@{}c@{}}$\Delta$MSE\\ (cn-tw)\end{tabular}} & \multicolumn{1}{l}{} & \textbf{tw} & \textbf{cn} & \textbf{\begin{tabular}[c]{@{}c@{}}$\Delta$MSE\\ (cn-tw)\end{tabular}} & \multicolumn{1}{l}{} & \textbf{tw} & \textbf{cn} & \textbf{\begin{tabular}[c]{@{}c@{}}$\Delta$MSE\\ (cn-tw)\end{tabular}} & \multicolumn{1}{l}{} \\ \midrule
 & \textbf{GPT-4o} & 3.563 & 3.769 & \cellcolor[HTML]{DBF3F4}0.206\hspace{4mm} & *** & 4.091 & 3.385 & \cellcolor[HTML]{EEA7A1}-0.706\hspace{4mm} & *** & 4.347 & 3.561 & \cellcolor[HTML]{EC9D96}-0.786\hspace{4mm} & *** \\
 & \textbf{Llama3 8b} & 2.187 & 2.268 & \cellcolor[HTML]{F1FAFB}0.082\hspace{4mm} & \multicolumn{1}{l}{} & 2.999 & 2.801 & \cellcolor[HTML]{FAE6E4}-0.199\hspace{4mm} & *** & 3.377 & 3.016 & \cellcolor[HTML]{F6D2CF}-0.361\hspace{4mm} & *** \\
 & \textbf{Llama3 70b} & 1.732 & 1.626 & \cellcolor[HTML]{FCF1F0}-0.107\hspace{4mm} & ** & 2.977 & 2.534 & \cellcolor[HTML]{F4C8C4}-0.443\hspace{4mm} & *** & 3.006 & 2.605 & \cellcolor[HTML]{F5CDC9}-0.401\hspace{4mm} & *** \\
 & \textbf{Llama3 405b} & 2.782 & 2.635 & \cellcolor[HTML]{FBECEB}-0.147\hspace{4mm} & \multicolumn{1}{l}{} & 4.624 & 3.685 & \cellcolor[HTML]{E88A82}-0.939\hspace{4mm} & *** & 4.620 & 3.740 & \cellcolor[HTML]{EA918A}-0.880\hspace{4mm} & *** \\
 & \textbf{Gemma2 9b} & 3.026 & 3.164 & \cellcolor[HTML]{E7F7F8}0.138\hspace{4mm} & * & 4.483 & 3.828 & \cellcolor[HTML]{EFADA8}-0.655\hspace{4mm} & *** & 4.928 & 4.131 & \cellcolor[HTML]{EC9C95}-0.797\hspace{4mm} & *** \\
\multirow{-6}{*}{\textbf{\begin{tabular}[c]{@{}c@{}}Short\\ (1-49)\end{tabular}}} & \textbf{Gemma2 27b} & 2.945 & 3.028 & \cellcolor[HTML]{F1FAFB}0.083\hspace{4mm} & \multicolumn{1}{l}{} & 4.888 & 4.191 & \cellcolor[HTML]{EEA8A2}-0.697\hspace{4mm} & *** & 4.944 & 4.250 & \cellcolor[HTML]{EEA9A3}-0.693\hspace{4mm} & *** \\ \midrule
 & \textbf{GPT-4o} & 1.846 & 1.577 & \cellcolor[HTML]{F8DDDB}-0.269\hspace{4mm} & *** & 1.834 & 1.57 & \cellcolor[HTML]{F8DEDC}-0.264\hspace{4mm} & *** & 2.070 & 1.743 & \cellcolor[HTML]{F7D6D3}-0.327\hspace{4mm} & *** \\
 & \textbf{Llama3 8b} & 1.674 & 1.548 & \cellcolor[HTML]{FBEFEE}-0.127\hspace{4mm} & *** & 2.046 & 1.895 & \cellcolor[HTML]{FBECEA}-0.152\hspace{4mm} & *** & 2.127 & 1.906 & \cellcolor[HTML]{F9E3E1}-0.220\hspace{4mm} & *** \\
 & \textbf{Llama3 70b} & 1.473 & 1.302 & \cellcolor[HTML]{FAE9E8}-0.171\hspace{4mm} & *** & 1.534 & 1.406 & \cellcolor[HTML]{FBEFEE}-0.128\hspace{4mm} & ** & 1.671 & 1.495 & \cellcolor[HTML]{FAE9E7}-0.176\hspace{4mm} & *** \\
 & \textbf{Llama3 405b} & 1.910 & 1.674 & \cellcolor[HTML]{F9E1DF}-0.236\hspace{4mm} & *** & 1.909 & 1.766 & \cellcolor[HTML]{FBEDEC}-0.143\hspace{4mm} & * & 2.085 & 1.892 & \cellcolor[HTML]{FAE6E5}-0.194\hspace{4mm} & ** \\
 & \textbf{Gemma2 9b} & 2.479 & 2.337 & \cellcolor[HTML]{FBEDEC}-0.142\hspace{4mm} & ** & 2.199 & 2.024 & \cellcolor[HTML]{FAE9E7}-0.175\hspace{4mm} & *** & 2.511 & 2.294 & \cellcolor[HTML]{F9E4E2}-0.217\hspace{4mm} & *** \\
\multirow{-6}{*}{\textbf{\begin{tabular}[c]{@{}c@{}}Long\\ (50+)\end{tabular}}} & \textbf{Gemma2 27b} & 2.703 & 2.519 & \cellcolor[HTML]{FAE8E6}-0.184\hspace{4mm} & *** & 2.680 & 2.500 & \cellcolor[HTML]{FAE8E7}-0.180\hspace{4mm} & *** & 2.649 & 2.496 & \cellcolor[HTML]{FBECEA}-0.153\hspace{4mm} & ** \\ \midrule
 & \textbf{GPT-4o} & 2.978 & 3.022 & \cellcolor[HTML]{F8FDFD}0.044\hspace{4mm} & \multicolumn{1}{l}{} & 3.323 & 2.767 & \cellcolor[HTML]{F1BAB5}-0.555\hspace{4mm} & *** & 3.571 & 2.942 & \cellcolor[HTML]{F0B0AB}-0.630\hspace{4mm} & *** \\
 & \textbf{Llama3 8b} & 2.011 & 2.021 & \cellcolor[HTML]{FEFFFF}0.010\hspace{4mm} & \multicolumn{1}{l}{} & 2.672 & 2.490 & \cellcolor[HTML]{FAE8E6}-0.182\hspace{4mm} & *** & 2.948 & 2.635 & \cellcolor[HTML]{F7D8D5}-0.313\hspace{4mm} & *** \\
 & \textbf{Llama3 70b} & 1.644 & 1.515 & \cellcolor[HTML]{FBEEED}-0.129\hspace{4mm} & *** & 2.486 & 2.150 & \cellcolor[HTML]{F7D5D2}-0.335\hspace{4mm} & *** & 2.551 & 2.227 & \cellcolor[HTML]{F7D6D4}-0.324\hspace{4mm} & *** \\
 & \textbf{Llama3 405b} & 2.483 & 2.306 & \cellcolor[HTML]{FAE9E7}-0.177\hspace{4mm} & *** & 3.695 & 3.028 & \cellcolor[HTML]{EFACA6}-0.667\hspace{4mm} & *** & 3.752 & 3.107 & \cellcolor[HTML]{EFAEA9}-0.645\hspace{4mm} & *** \\
 & \textbf{Gemma2 9b} & 2.840 & 2.882 & \cellcolor[HTML]{F8FDFD}0.043\hspace{4mm} & \multicolumn{1}{l}{} & 3.705 & 3.213 & \cellcolor[HTML]{F3C2BD}-0.491\hspace{4mm} & *** & 4.105 & 3.505 & \cellcolor[HTML]{F0B4AF}-0.600\hspace{4mm} & *** \\
\multirow{-6}{*}{\textbf{Overall}} & \textbf{Gemma2 27b} & 2.863 & 2.855 & \cellcolor[HTML]{FEFEFD}-0.008\hspace{4mm} & \multicolumn{1}{l}{} & 4.136 & 3.615 & \cellcolor[HTML]{F2BEB9}-0.521\hspace{4mm} & *** & 4.162 & 3.653 & \cellcolor[HTML]{F2BFBB}-0.509\hspace{4mm} & *** \\ \bottomrule
\end{tabular}
\vspace{-.5pc}
\caption{Mean squared error (MSE~$\downarrow$) by length for GPT-4o, Llama3 (8b, 70b, 405b), and Gemma2 (9b, 27b) models. Statistical significance notations and color coding follow the same conventions as in Table~\ref{tab:new-mse-results}.}
\vspace{-1pc}
\label{tab:new-mse-results}
\end{table*}

\section{Constructing a Contextually-Aligned Review Dataset for Language Varieties}

\vspace{-.5pc}
\paragraph{Data.}
We constructed a dataset of hotel reviews sourced from \texttt{Booking.com},\footnote{Data processing code: \href{https://github.com/Crowd-AI-Lab/Contextually-Aligned-Online-Reviews}{https://github.com/Crowd-AI-Lab/Contextually-Aligned-Online-Reviews}} which has been used in prior research studies~\cite{ALDERIGHI2022769,barnes-etal-2018-multibooked}.
This dataset consists of 4,447,853 reviews labeled by the platform as written in Chinese.
The reviews cover 149,879 hotels located in Japan, Mainland China, South Korea, Taiwan, Thailand, and Vietnam, and were collected from August 2021 to August 2024. 
These locations were selected to ensure a substantial volume of data, as they are popular destinations for Mandarin-speaking travelers.
Each review comprises three main components: the review title, positive feedback, and negative feedback.
Additionally, it includes review ratings (ranging from 1 to 10 stars) and metadata such as hotel ID, posting time, and more (see \Cref{app:booking-data-sample} for an actual sample). 
Booking.com claims to invest significant effort in ensuring that reviews are posted by real users and in maintaining review quality. 
We included only non-empty reviews, meaning reviewers provided input in at least one of the following: 
the review title, positive feedback, or negative feedback. 
In total, we collected 1,513,056 reviews written in Chinese.







\vspace{-.6pc}
\subsection{Contextually Aligning Reviews}
We used users' self-specified ``nationality/region'' labels from Booking.com to determine the reviews' language varieties. 
In total, we collected 1,403,669 reviews written in \twChinese and \cnChinese, where 95.591\% of them come from \twChinese users.
To ensure a balanced representation between \textbf{\twChinese (TW)} and \textbf{\cnChinese (CN)} reviews, we paired them based on the following criteria:

\vspace{-.8pc}

\begin{itemize}[leftmargin=*]
\item
\textbf{Same hotel for both reviews:} 
Both reviews in each pair are from the same hotel, ensuring that the reviewers are commenting on similar scenarios or objects---the hotel itself.

\vspace{-.5pc}

\item
\textbf{Similar ratings for both reviews:} 
To form comparable pairs with similar sentiments, we used a 3-class rating scheme (1-3 as negative, 4-7 as neutral, and 8-10 as positive) and paired reviews based on this classification. 
This approach maximizes the number of review pairs while maintaining comparable sentiment.

\vspace{-.5pc}

\item
\textbf{Similar text length for both reviews:} 
To ensure paired reviews have similar text lengths, we grouped reviews into 10-token bins before pairing and required both reviews in each pair to fall within the same length bin.
Reviews longer than 500 tokens were excluded (see \Cref{app:length-exp}.)

\end{itemize}

\vspace{-.5pc}

The final dataset contained 22,918 review pairs, each with one TW and one CN user review.


\subsection{Data Quality Validation\label{sec:data-quality-validation}}
Five native speakers of \twChinese reviewed 200 random \twChinese reviews; the same process applied to \cnChinese.
The focus was on two key aspects: {\em (i)} \textbf{writing quality} and {\em (ii)} \textbf{content-rating agreement}, evaluated on a 5-point Likert scale (see Appendix~\ref{app:human-validation}.) 
Each participant was paid \$10.
As a result, for the writing quality ratings, the TW group had a mean of 4.18 (SD=0.44), and the CN group had a mean of 3.94 (SD=0.49). 
Regarding the rating-content agreement, the TW group had a mean of 4.00 (SD=0.46), and the CN group had a mean of 3.56 (SD=0.55).

\section{Experimental Results\label{sec:experiment}}




To examine biases from review structure, we tested three settings:
{\em (i)} \textbf{Structured review} retains the original format with title, positive, and negative feedback.
{\em (ii)} \textbf{Plain review} concatenates all elements into a single paragraph.
{\em (iii)} \textbf{Shuffled review} includes all elements but in random order.
For the analysis, we excluded pairs that lacked complete predictions or received predictions that did not follow the specified format (see \Cref{appendix:valid-sample-count}).
Once the contextually aligned dataset was constructed and available, we tested it using six LLMs: GPT-4o, Llama3 (8b, 70b, 405b), and Gemma2 (9b, 27b). 
The task involved predicting a rating score (from 1 to 10, where 1 is the worst and 10 is the best) based on the review content.
The prompt (\Cref{app:prompts}) includes the task description, the review content, and the prediction scale (1-10). 
\Cref{tab:new-acc-results} and \Cref{tab:new-mse-results} show the prediction accuracy (Acc) and mean squared error (MSE) across models and settings (see \Cref{appendix:valid-sample-count} for valid prediction counts.)

\paragraph{LLMs performed significantly worse in \twChinese compared to \cnChinese.}
Among all 54 experiments with different models and prompt settings, 38 of them had significant group differences in accuracy (Table~\ref{tab:new-acc-results}), and 47 had significant group differences in MSE (Table~\ref{tab:new-mse-results}).
Among all significant accuracy differences, LLMs made less accurate sentiment predictions toward \twChinese users (36 out of 38 in Acc, and 45 out of 47 in MSE).


\paragraph{When the reviews' structures are disrupted, the performance gap increases.}
\Cref{tab:new-acc-results} and \Cref{tab:new-mse-results} show that 
structured input reduces performance gaps and generally improves model performance.
Without knowing the structure inside reviews (\ie, plain or shuffled cases), bias toward \twChinese and \cnChinese increases.

\paragraph{Shorter reviews tend to produce larger MSE gaps.}
Our pilot study (\Cref{app:length-exp}) found that shorter texts may lack information and often affect model performance and behavior.
We thus categorized our dataset into two groups based on review's text length: 
short (1-49 Chinese characters) and long (50+ Chinese characters). 
\Cref{tab:new-mse-results} shows that the MSE gap between \twChinese and \cnChinese widens in the short text group (also see \Cref{fig:length-exp} in \Cref{app:length-exp}), while this trend is less clear for Acc (\Cref{tab:new-acc-results}).


\begin{table}[t]
\centering
\footnotesize
\addtolength{\tabcolsep}{-1.2mm}
\begin{tabular}{@{}ccccr@{\kern-4mm}lccr@{\kern-4mm}l@{}}
\toprule
\multicolumn{1}{l}{} & \multicolumn{1}{l}{} & \multicolumn{4}{c}{\textbf{Acc$\uparrow$}} & \multicolumn{4}{c}{\textbf{MSE$\downarrow$}} \\ \cmidrule(lr){3-6} \cmidrule(lr){7-10}
\multicolumn{1}{l}{} & \textbf{Ori.} & \textbf{tw} & \textbf{cn} & \multicolumn{2}{c}{\textbf{\begin{tabular}[c]{@{}c@{}}$\Delta$Acc\\ (cn-tw)\end{tabular}}} & \textbf{tw} & \textbf{cn} & \multicolumn{2}{c}{\textbf{\begin{tabular}[c]{@{}c@{}}$\Delta$MSE\\ (cn-tw)\end{tabular}}} \\ \midrule
 & \textbf{tw} & 29.60 & 30.20 & \cellcolor[HTML]{FCF0EF}0.60\hspace{4mm} & * & 2.985 & 2.036 & \cellcolor[HTML]{E88981}-0.948\hspace{4mm} & *** \\
\multirow{-2}{*}{\textbf{stru.}} & \textbf{cn} & 30.31 & 31.16 & \cellcolor[HTML]{FBE9E8}0.85\hspace{4mm} & ** & 1.969 & 3.026 & \cellcolor[HTML]{46BDC6}1.056\hspace{4mm} & *** \\ \midrule
 & \textbf{tw} & 22.26 & 23.06 & \cellcolor[HTML]{FCEBE9}0.80\hspace{4mm} & *** & 3.262 & 2.577 & \cellcolor[HTML]{EEA9A4}-0.686\hspace{4mm} & *** \\
\multirow{-2}{*}{\textbf{plain}} & \textbf{cn} & 24.03 & 25.02 & \cellcolor[HTML]{FBE6E4}0.99\hspace{4mm} & *** & 2.267 & 2.727 & \cellcolor[HTML]{AFE3E7}0.460\hspace{4mm} & *** \\ \midrule
 & \textbf{tw} & 21.40 & 22.10 & \cellcolor[HTML]{FCEDEC}0.70\hspace{4mm} & *** & 3.489 & 2.688 & \cellcolor[HTML]{EC9B94}-0.802\hspace{4mm} & *** \\
\multirow{-2}{*}{\textbf{shuf.}} & \textbf{cn} & 23.48 & 24.01 & \cellcolor[HTML]{FDF2F1}0.53\hspace{4mm} & ** & 2.393 & 2.901 & \cellcolor[HTML]{A6E0E4}0.508\hspace{4mm} & *** \\ \bottomrule
\end{tabular}
\addtolength{\tabcolsep}{+1.2mm}
\vspace{-.7pc}
\caption{GPT-4o performance on original (Ori.) and machine-translated texts. TW-to-CN translation improved Acc and MSE; CN-to-TW showed mixed results.
Statistical significance notations and color coding follow the same conventions as in Table~\ref{tab:new-mse-results}.
}
\vspace{-1.7pc}
\label{tab:translation-result}
\end{table}

\subsection{Can We Just Use Machine Translation?}
A natural question is whether we could use machine translation to convert \twChinese to \cnChinese, and vice versa, to create a paired dataset for benchmarking. 
To explore this, 
we translated all texts to their opposite version (\twChinese to \cnChinese, or vice versa) using the Google Translate API.
We then conducted sentiment analysis experiments using GPT-4o, comparing each original sample with its translated version (\eg, [a review in TW, its translation into CN].)
The results (\Cref{tab:translation-result}) show an \textbf{asymmetry between the two translation directions}.
Translating \twChinese data to \cnChinese increased accuracy and decreased MSE (\Cref{tab:translation-result}'s 1st, 3rd, and 5th rows).
However, translating \cnChinese to \twChinese produced mixed results: it decreased accuracy but improved MSE.
These results suggest that while using machine translation to create review pairs between language varieties is technically feasible, it can introduce an additional layer of bias, as machine translation itself is a language technology that is not immune from biases across language varieties.
In our case, machine translation might be better at \twChinese to \cnChinese than the other way around \cite{kantharuban2023quantifying}. 
Furthermore, mature machine translation systems for specific language varieties are not always readily available~\cite{multivaluebench,kumar-etal-2021-machine}.
\section{Examining Confounding Variables}

\paragraph{Could the performance gap be due to \cnChinese reviews having better \ul{writing quality} or better \ul{alignment between content and ratings}?}
\textit{Rationale:} Better writing quality or better content-rating alignment could make it easier for LLMs to predict ratings.
\textit{Analysis \& Findings:} \textbf{No.} 
Our human validation (Section~\ref{sec:data-quality-validation}) shows that \cnChinese reviews had slightly worse writing quality and content-rating alignment.

\paragraph{Could the performance gap be due to more \ul{code-mixed usage} in \twChinese?}
\textit{Rationale:} NLP models often struggle with code-mixed data~\cite{zhang-etal-2023-multilingual, ochieng2024beyond}. 
\textit{Analysis \& Findings:}
\textbf{No.}
The \cnChinese reviews contain more mixed-language input (30.99\%) than the \twChinese reviews (25.26\%, see Appendix~\ref{appendix:language-analysis} and Table~\ref{tab:language-distribution}).

\paragraph{Could the performance gap be due to \cnChinese \ul{users} systematically \ul{giving higher scores}, which align better with LLM-generated scores?}
\textit{Rationale:} LLMs tend to assign higher scores~\cite{stureborg2024large,kobayashi-etal-2024-large,golchin-etal-2025-grading}.
\textit{Analysis \& Findings:}
\textbf{Unlikely.}
In our dataset, \twChinese and \cnChinese reviews show no significant difference in scores (\textit{t}(22917) = .160, \textit{p} = .873).

\paragraph{Are \cnChinese reviews \ul{easier for humans to guess ratings}?}
\textit{Rationale:} Human performance is sometimes used as an indicator of a task's difficulty for LLMs~\cite{sakamoto-etal-2025-development,ding2024easyhardbench}.
\textit{Analysis \& Findings:}
\textbf{Plausible.}
We conducted a user study with 10 participants (5 native speakers from each variety) who reviewed 50 random CN-TW review pairs (100 total reviews) and predicted their rating scores.
Participants performed significantly better at predicting ratings for reviews in \cnChinese.
After excluding two TW native speakers whose accuracy was more than two standard deviations below the mean, 6 out of the 8 participants had better accuracy on CN reviews than TW reviews, and 7 had better (lower) MSE on CN reviews than TW reviews (see \Cref{app:human-prediction} for more details).

These results should be interpreted with caution.
Unlike question-answering, predicting hundreds of review scores from content is not a typical human task, and most NLP papers on sentiment analysis do not compare model performance to human performance.
Thus, it is unclear whether human performance gaps in such tasks reliably indicate task difficulty for LLMs, especially given the small differences between the two varieties.
Additionally, our participants may not represent the average Mandarin speaker's ability in sentiment analysis, as the two participants performed notably poorly. 
Finally, despite our efforts to examine confounding variables such as text length, code-mixing, and writing quality, we still \textbf{lack a clear understanding of what causes the observed LLMs' performance gaps across language varieties}.






\section{Discussion}


\paragraph{Do users who self-label as being from Taiwan always use \twChinese?}
In this study, we use users' self-reported nationality/region to infer whether they are speakers of \twChinese or \cnChinese. 
The convention is that \twChinese employs traditional Chinese characters, while \cnChinese uses simplified characters. 
However, analysis using predefined character sets revealed that 30.99\% of samples in the CN group contained characters beyond simplified Chinese, and 25.26\% of samples in TW group included characters not limited to traditional Chinese. 
This suggests that the relationship between self-reported nationality/region, language variety, and character usage is more complex in real-world data.
In \Cref{appendix:language-analysis}, \Cref{tab:language-distribution} shows the distribution of Chinese script variants among users.

\section{Conclusion and Future Work}

This paper introduces a cost-effective method for benchmarking model performance across language varieties using international online reviews from similar contexts.
To validate this, we built a contextually aligned dataset of \twChinese and \cnChinese reviews and tested six LLMs on sentiment analysis, finding that LLMs consistently underperform in \twChinese. 
We aim to extend this approach to more language varieties, with the ultimate goal of creating LLMs that perform equally well across them.


\section{Limitations}
As the study that is among the first to benchmark LLMs' performance across language varieties using contextually aligned data, this study and the data pairing method we introduced have several limitations. 

\begin{itemize}
    \item 

The first limitation is that, despite the contextual alignment, unknown confounding factors might contribute to performance gaps. 
This is an inherent challenge when using user-generated data in the wild for apple-to-apple comparisons, as controlling all variables is almost impossible.
Relaxing strict semantic alignment between paired text items inevitably introduces confounding variables.
We believe that this trade-off is worth exploring because it enables researchers to compare model behaviors across language varieties in new ways. 

\item 
Another limitation relates to the input prompts, which are code-mixed. Previous studies found that LLMs might still have deficits in dealing with cultural context and code-mixing input~\cite{ochieng2024beyond}. 
We used English for instruction to exclude potential biases introduced if it is prompted in Chinese, regardless of its variety. 
However, such a setup may introduce additional confusion for LLMs to process, leading to lower performance results. The usage of English prompts regarding non-English tasks, or code-switching prompts, requires thorough studies to better investigate LLMs' capability of multilingualism and awareness of language and cultural diversity.

\item 
A third limitation concerns our machine translation-based analysis.
We recognize that the observed performance differences when translating between \twChinese and \cnChinese
may arise from a combination of morphosyntactic variations, script differences,
and normalization of non-Chinese script elements.
More importantly, while MT-based approaches are technically feasible,
they can introduce additional biases, as MT systems themselves exhibit performance disparities across language varieties.
Further analyses are required to better isolate and address these compounding factors.

\end{itemize}

\section{Ethics Statement}
We assess that the general risks and ethical concerns of our work are no greater than those involved in using user-generated reviews to test sentiment analysis models.

\section*{Acknowledgement}
We thank the anonymous reviewers for their feedback and the participants for their contributions to our human studies. 
This work was partially supported by the 2024-2025 Seed Grant from the College of Information Sciences and Technology at Pennsylvania State University. 
We also acknowledge Dr. Janet G. van Hell, Co-PI of the seed grant, for her support and valuable input. 
Additionally, this work was partially supported by the National Science and Technology Council (NSTC), Taiwan, under the project ``\textit{Taiwan's 113th Year Endeavoring in the Promotion of a Trustworthy Generative AI Large Language Model and the Cultivation of Literacy Capabilities (Trustworthy AI Dialog Engine, TAIDE)}.''

\bibliography{bib/custom}

\appendix

\section{Booking.com Data\label{app:booking-data-sample}}
\Cref{tab:sample data} shows a sample of the collected Booking.com review.

\newcolumntype{Y}{>{\raggedright\arraybackslash}X}

\begin{table*}[htbp]
    \centering \small

    \begin{tabularx}{\textwidth}{@{}lY@{}}
        \toprule
        \textbf{Field} & \textbf{Value} \\
        \midrule
        \texttt{hotel\_\_booking\_id} & \texttt{311092} \\
        \texttt{hotel\_\_ufi} & \texttt{-240213} \\
        \texttt{user} & --------- (\textit{Removed the user identity}) \\
        \texttt{user\_nationality} & \texttt{tw} \\
        \texttt{room\_type} & \begin{CJK*}{UTF8}{bsmi}雙床房－附加床－禁煙\end{CJK*} \\
        & \textit{(English Translation: Twin Room - Extra Bed - Non-Smoking)} \\
        \texttt{checking\_date} & \texttt{2023-04-23} \\
        \texttt{checkout\_date} & \texttt{2023-04-26} \\
        \texttt{length\_of\_stay} & \texttt{3} \\
        \texttt{guest\_type} & \texttt{null} \\
        \texttt{score} & \texttt{10.0} \\
        \texttt{review\_title} & \texttt{null} \\
        \texttt{positive\_review} & \begin{CJK*}{UTF8}{bsmi}櫃檯很友善，有事情都很熱心協助，環境乾淨整潔，住的很舒適，還貼心附上各種充電頭，超級滿意！\end{CJK*} \\
        & \textit{(English Translation: The front desk is very friendly and helpful. The environment is clean and tidy. The stay was comfortable. They thoughtfully provided various charging heads. Super satisfied!)} \\
        \texttt{negative\_review} & \texttt{null} \\
        \texttt{hotel\_response} & \texttt{null} \\
        \texttt{review\_time} & \texttt{2023-05-15 10:55:59+00:00} \\
        \texttt{created} & \texttt{2024-08-18 07:11:29.971276+00:00} \\
        \bottomrule
        \multicolumn{2}{l}{{\footnotesize\textit{*Note: English translations in italics are provided for readability and are not part of the actual data.}}}
    \end{tabularx}
    \caption{Sample data entry from the collected Booking.com. There are three review components: review\_title, positive\_review, and negative\_review.}
    \label{tab:sample data}
\end{table*}

\section{Human Validation}
\subsection{Questions for data quality validation}\label{app:human-validation}
We used the following two questions in the human evaluation to assess data quality. 
For each part of the study, participants were shown both the English text and its translation, either into \twChinese or \cnChinese, depending on the context.

\begin{enumerate}
    \item The review (including the title, positive, and negative sections) is easy to read, and the writing quality is comparable to online reviews written by native speakers, based on my experience.

    \begin{itemize}[leftmargin=*]
        \item \begin{CJK*}{UTF8}{bsmi} \twChinese:
        根據我的經驗，這篇評論（包括標題、優點和缺點部分）很容易閱讀，且寫作品質與母語使用者撰寫的網路評論相當。\end{CJK*}
    
        \item \begin{CJK*}{UTF8}{gbsn} \cnChinese:
        根据我的经验，这篇评论（包括标题、优点和缺点部分）很容易阅读，而且写作质量与母语者撰写的网络评论相当。\end{CJK*}
    
    \end{itemize}
    
    \item The score (1-10, 1 is the worst, 10 is the best) assigned to this review accurately reflects the content of the review.

    \begin{itemize}[leftmargin=*]
        \item \begin{CJK*}{UTF8}{bsmi} \twChinese:
        這篇評論的分數（1-10，1是最差，10是最好）準確反映了評論的內容。
        \end{CJK*}
        
        \item \begin{CJK*}{UTF8}{gbsn} \cnChinese:
        这篇评论的评分（1-10，1是最差，10是最好）准确反映了评论的内容。
        \end{CJK*}
    \end{itemize}
    
\end{enumerate}
\subsection{Score prediction}\label{app:human-prediction}\zixin{new subsection.}

We used the following questions to further investigate potential content differences in review pairs, which can further lead to gaps in LLMs' performance differences. In this study, participants were asked to rate 1) the readability of the review, 2) the overall nativeness of the review, and 3) the score of the review. For the convenience of reading, all reviews were converted into either traditional or simplified Chinese characters so that all participants could process them in the writing style of their native language variety. Both English and its translation, in either \cnChinese or \twChinese based on the participants' language background, were provided in the instruction.
\begin{enumerate}
    \item Readability (1-5), where: 1 = The writing doesn't contain any literal information; 3 = The writing requires additional effort to process/comprehend; 5 = The writing is fluent and clear in terms of content delivery
    \begin{itemize}[leftmargin=*]
        \item \begin{CJK*}{UTF8}{bsmi}\twChinese: 評論可讀性(1-5分)，其中：1分表示評論不具備可讀性，或其語句無任何實際意義；3分表示評論存在語句不通的情況，且該情況會導致歧義或理解困難；5分表示評論語句通順，表達連貫，語義明確且清晰。
        \end{CJK*}
        \item \begin{CJK*}{UTF8}{gbsn}\cnChinese: 评论可读性(1-5分)，其中：1分表示评论不具备可读性，或其语句无任何实质意义；3分表示评论存在语句不通的情况或语病，且该情况会影响阅读或理解；5分表示评论语句通顺，表达连贯，语义明确且清晰。
        \end{CJK*}
    \end{itemize}
    
    \item Nativeness - the review is generated by: 1. a less proficient non-native Chinese speaker; 2. a highly proficient non-native Chinese speaker or a native Chinese speaker; 3. machine translation from another language; or 4. not sure/inconclusive
    \begin{itemize}[leftmargin=*]
        \item \begin{CJK*}{UTF8}{bsmi} \twChinese: 你覺得該評論可能出自：1. 低水平的中文非母語者；2. 高水平的中文非母語者或中文母語者；3. 來自其他語言的機器翻譯；4. 不確定/無法判斷。
        \end{CJK*}
        \item \begin{CJK*}{UTF8}{gbsn} \cnChinese: 你觉得该评论可能出自：1. 低水平中文非母语者；2. 高水平中文非母语者或中文母语者；3. 来自其他语言的机器翻译；4. 不确定/无法判断。
        \end{CJK*}
    \end{itemize}

    \item Score Rating (1-10, 1 is the lowest, 10 is the highest)
    \begin{itemize}[leftmargin=*]
        \item \begin{CJK*}{UTF8}{bsmi} \twChinese: 旅館評分 (1-10，1為最差，10為最好)
        \end{CJK*}
        \item \begin{CJK*}{UTF8}{gbsn} \cnChinese: 酒店评分 (1-10，1为最差，10为最好)
        \end{CJK*}
    \end{itemize}
\end{enumerate}

We further excluded two participants' responses due to the lack of score agreement against other participants and their significantly lower performance in prediction accuracy. Among the other 8 participants, there are no significant differences in score predictions among the data pairs, indicating raters have no biases in reading and understanding reviews from either group of speakers/writers. However, results showed statistical significance in both Accuracy (37.00\% vs. 28.75\%, \textit{p}=.016) and MSE (2.795 vs. 3.510, \textit{p}=.036), showing that native speakers might have more difficulties in correctly guessing the review scores for reviews in \twChinese. 

\section{Prompts\label{app:prompts}}

The following prompt is used for the structured condition.
\begin{quote}
    \small
    \textbf{System} \\
    \texttt{You are a grading assistant for hotel reviews}

    \textbf{User} \\
    \texttt{
        The following is a hotel review from a user. Based on the title, positive feedback, and negative feedback provided below, give an overall score from 1 to 10, where 1 is the worst and 10 is the best. DO NOT include any words in your output, just provide the number. \\ \\
        Title: [title] \\
        Positive Feedback: [positive\_review] \\
        Negative Feedback: [negative\_review] \\
        Overall Score (1-10):
    }
\end{quote}

The following prompt is used for both the plain and shuffled conditions.
\begin{quote}
    \small
    \textbf{System} \\
    \texttt{You are a grading assistant for hotel reviews}

    \textbf{User} \\
    \texttt{
        The following is a hotel review from a user. Based on the input review below, give an overall score from 1 to 10, where 1 is the worst and 10 is the best. DO NOT include any words in your output, just provide the number. \\ \\
        input: [text] \\
        Overall Score (1-10):
    }
\end{quote}

For LLMs that don't have a system role setting (\eg Gemma2), the system instruction is removed from the prompts.

\begin{table*}[]
\centering \small
\addtolength{\tabcolsep}{-1.2mm}
\begin{tabular}{@{}cccccccccc@{}}
\toprule
\multirow{2}{*}{\textbf{model}} & \multicolumn{3}{c}{\textbf{All}} & \multicolumn{3}{c}{\textbf{Short}} & \multicolumn{3}{c}{\textbf{Long}} \\ \cmidrule(lr){2-4} \cmidrule(lr){5-7} \cmidrule(lr){8-10}
 & \textbf{plain} & \textbf{shuffled} & \textbf{structured} & \textbf{plain} & \textbf{shuffled} & \textbf{structured} & \textbf{plain} & \textbf{shuffled} & \textbf{structured} \\ \midrule
\textbf{GPT-4o} & 45,828 & 45,830 & 45,836 & 45,828 & 45,830 & 45,836 & 45,836 & 45,836 & 45,836 \\
\textbf{LLaMA-3.1 8B} & 45,668 & 45,707 & 45,697 & 45,694 & 45,726 & 45,712 & 45,810 & 45,817 & 45,821 \\
\textbf{LLaMA-3.1 70B} & 45,835 & 45,835 & 45,834 & 45,835 & 45,835 & 45,834 & 45,836 & 45,836 & 45,836 \\
\textbf{LLaMA 3.1 405B} & 45,805 & 45,795 & 45,706 & 45,808 & 45,801 & 45,710 & 45,833 & 45,830 & 45,832 \\
\textbf{Gemma-2 9B} & 45,836 & 45,836 & 45,819 & 45,836 & 45,836 & 45,820 & 45,836 & 45,836 & 45,835 \\
\textbf{Gemma-2 27B} & 45,833 & 45,833 & 45,824 & 45,833 & 45,833 & 45,824 & 45,836 & 45,836 & 45,836 \\ \midrule
\textbf{GPT-4o+Translation} & 45,682 & 45,644 & 45,836 & - & - & - & - & - & - \\ \bottomrule
\end{tabular}
\addtolength{\tabcolsep}{+1.2mm}
\caption{Number of valid prediction samples in the study across different models and data configurations. }
\label{tab:valid-num-samples}
\end{table*}

\begin{table*}[]
\centering \small
\addtolength{\tabcolsep}{-1.2mm}
\sisetup{
  table-format=-4.0,
  group-separator={,},
  group-minimum-digits=4,
  print-zero-integer=true,
}
\begin{tabular}{@{}l*{9}{S}@{}}
\toprule
\multirow{2}{*}{\textbf{model}} & \multicolumn{3}{c}{\textbf{All}} & \multicolumn{3}{c}{\textbf{Short}} & \multicolumn{3}{c}{\textbf{Long}} \\ \cmidrule(lr){2-4} \cmidrule(lr){5-7} \cmidrule(lr){8-10}
 & \textbf{plain} & \textbf{shuffled} & \textbf{structured} & \textbf{plain} & \textbf{shuffled} & \textbf{structured} & \textbf{plain} & \textbf{shuffled} & \textbf{structured} \\ \midrule
\textbf{GPT-4o} & -8 & -6 & 0 & -8 & -6 & 0 & 0 & 0 & 0 \\
\textbf{LLaMA-3.1 8B} & -168 & -129 & -139 & -142 & -110 & -124 & -26 & -19 & -15  \\
\textbf{LLaMA-3.1 70B} & -1 & -1 & -2 & -1 & -1 & -2 & -0 & -0 & -0 \\
\textbf{LLaMA 3.1 405B} & -31 & -41 & -130 & -28 & -35 & -126 & -3 & -6 & -4 \\
\textbf{Gemma-2 9B} & 0 & 0 & -17 & 0 & 0 & -16 & 0 & 0 & -1 \\
\textbf{Gemma-2 27B} & -3 & -3 & -12 & -3 & -3 & -12 & 0 & 0 & 0 \\ \midrule
\textbf{GPT-4o+Translation} & -154 & -192 & 0 & {-} & {-} & {-} & {-} & {-} & {-} \\ \bottomrule
\end{tabular}
\addtolength{\tabcolsep}{+1.2mm}
\caption{Number of invalid predictions in the study across different models and data configurations. Negative values indicate the count of invalid samples. Results show that some models (e.g., Gemma-2 27B and LLaMA-3.1 8B) exhibit substantially higher numbers of invalid samples, particularly for structured data.}
\label{tab:missing-num-samples}
\end{table*}

\section{Distribution of Valid and Invalid Predictions\label{appendix:valid-sample-count}}
\Cref{tab:valid-num-samples} and \Cref{tab:missing-num-samples} present the numbers of valid and invalid predictions obtained from our experimental procedures.
Invalid predictions encompass instances where models deviated from the task requirements,
such as providing explanations instead of numerical outputs,
generating values outside the specified range of 1-10,
or failing to engage with the task altogether.
We only included pairs with completely valid data entries for the prediction analysis (\Cref{tab:new-acc-results} and \Cref{tab:new-mse-results}), referring to the smallest number of each model in \Cref{tab:valid-num-samples}.

\begin{figure*}
    \centering
    \includegraphics[width=1.0\linewidth]{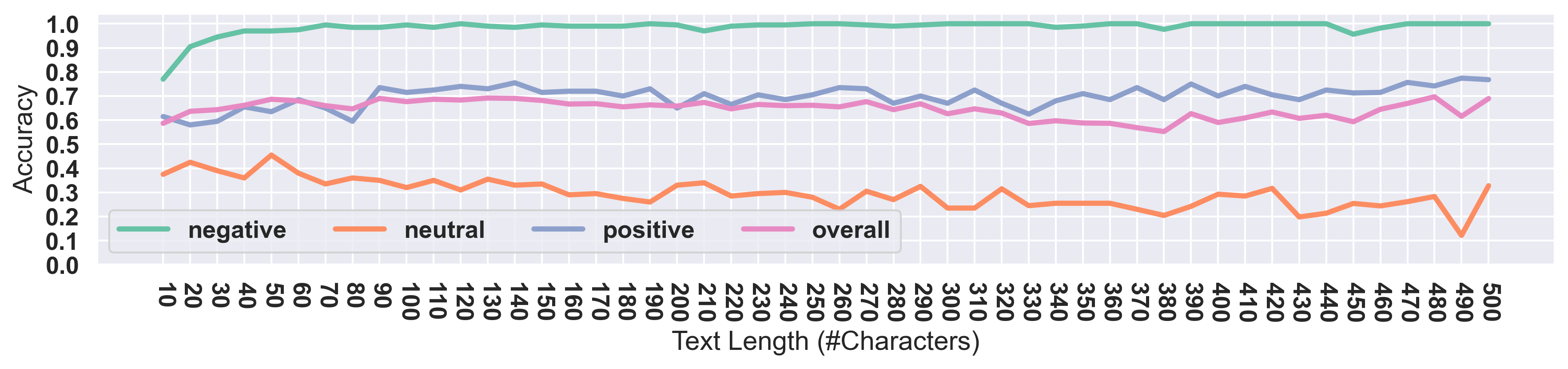}
    \includegraphics[width=1.0\linewidth]{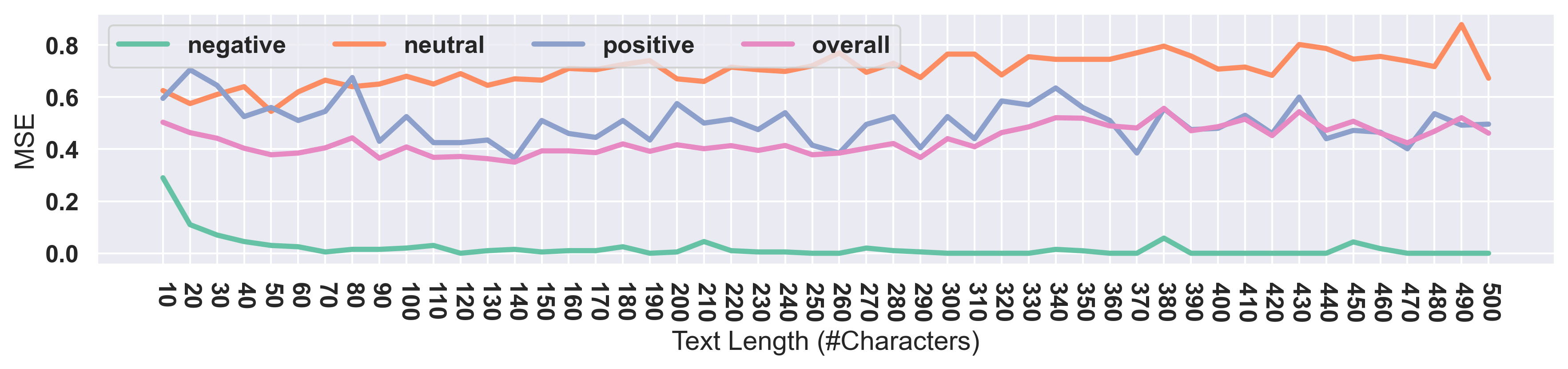}
    \caption{Impact of text length on sentiment classification performance. The top graph shows accuracy, and the bottom graph shows MSE for negative, neutral, positive, and overall sentiments across different text lengths (0-500 characters). While overall performance remains relatively stable, individual sentiment categories show varying levels of accuracy and error, particularly for shorter texts.}
    \label{fig:length-exp}
\end{figure*}

\section{Pilot Study on Impact of Text Length\label{app:length-exp}}
During our data exploration phase, we investigated whether short texts should be removed due to potentially insufficient information for accurate sentiment classification.
To address this, we conducted a pilot experiment to analyze the relationship between text length and model performance.

\paragraph{Data}
We used the initial Booking.com dataset, assigning sentiment labels based on review scores:
positive (8-10), neutral (4-7), and negative (1-3).
The input text was created by concatenating three review components:
\begin{quote}
    \small
    \texttt{[review-title]}\\
    \texttt{[positive-review]}\\
    \texttt{[negative-review]}
\end{quote}
We categorized the texts into 50 bins of 10 characters each, up to 500 characters in length.
For each bin, we selected a balanced set of 600 samples (200 per sentiment label) where possible.
It's worth noting that for texts longer than 290 characters, maintaining this balance became challenging due to insufficient samples.

\paragraph{Predictions}
We employed GPT-4o (\texttt{gpt-4o-2024-08-06}) to classify each sample into one of the three sentiment categories using the following prompt (without a system prompt):
\begin{quote}
    \small
    \textbf{User} \\
    \texttt{
        Predict the sentiment of the following text. Please answer one of the following label: (positive, negative, neutral). Do not reply anything like `The sentiment is...'. Do not replay with any explanation. Directly output the answer. \\ \\
        Text: [text]
    }
\end{quote}
Predictions outside the specified labels were excluded from the analysis (only one sample was removed in this experiment).

\paragraph{Results}
\Cref{fig:length-exp} illustrates the accuracy and MSE for each sentiment label and the overall performance across different text lengths.
While the overall performance remains relatively stable across text lengths, we observed variations in performance for individual sentiment labels.
This effect is particularly noticeable for negative sentiments in shorter texts.
Our findings indicate that text length does influence model performance, though not to the extent of completely compromising the model's ability to classify sentiments.
Based on these results, we decided against filtering samples based on text length.
Instead, we report scores for different text length groups (short: 1-49 and long: 50+) to provide a comprehensive view of the model's performance across text lengths.

\begin{figure*}
    \centering
    \includegraphics[width=0.48\linewidth]{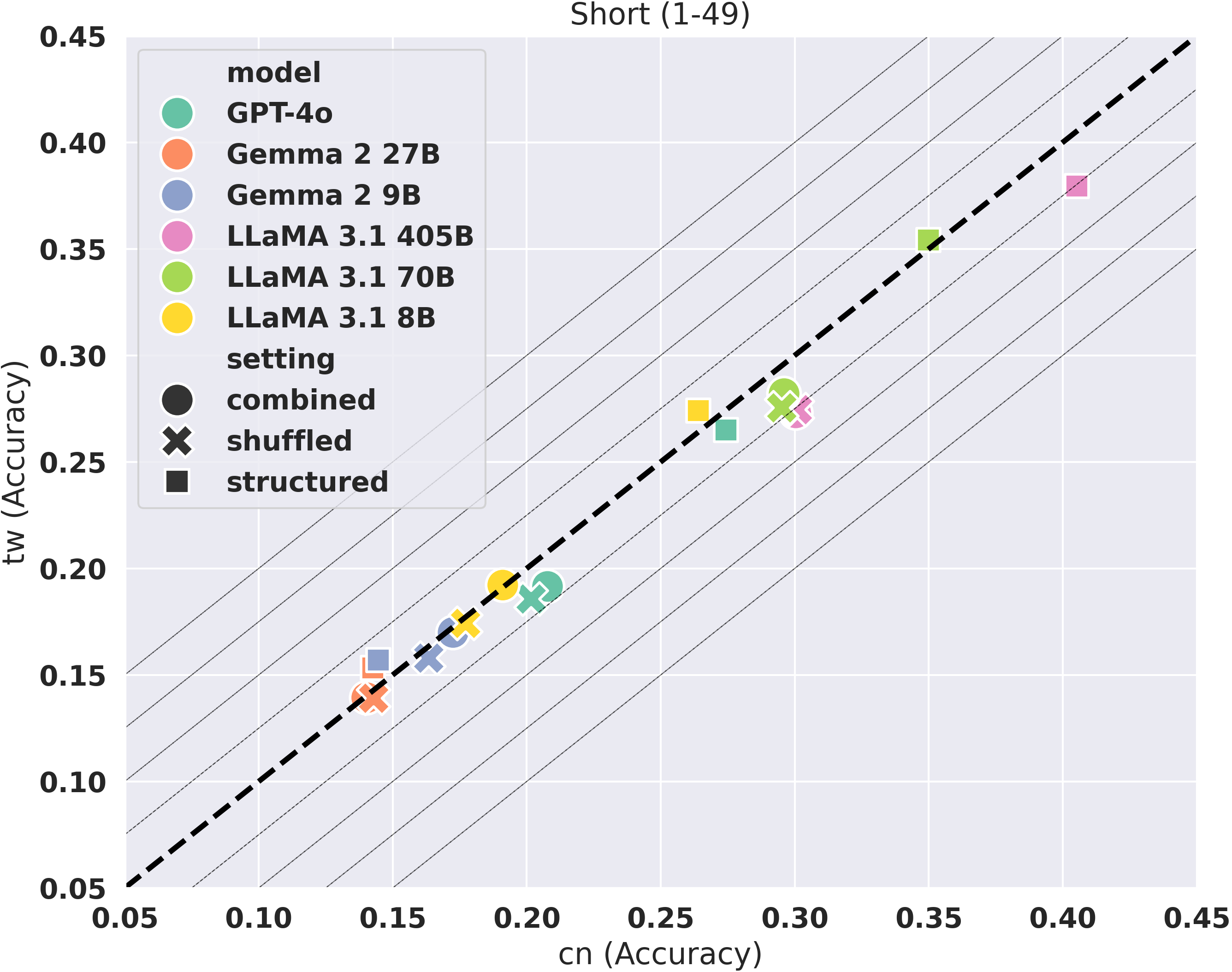}
    \hfill
    \includegraphics[width=0.48\linewidth]{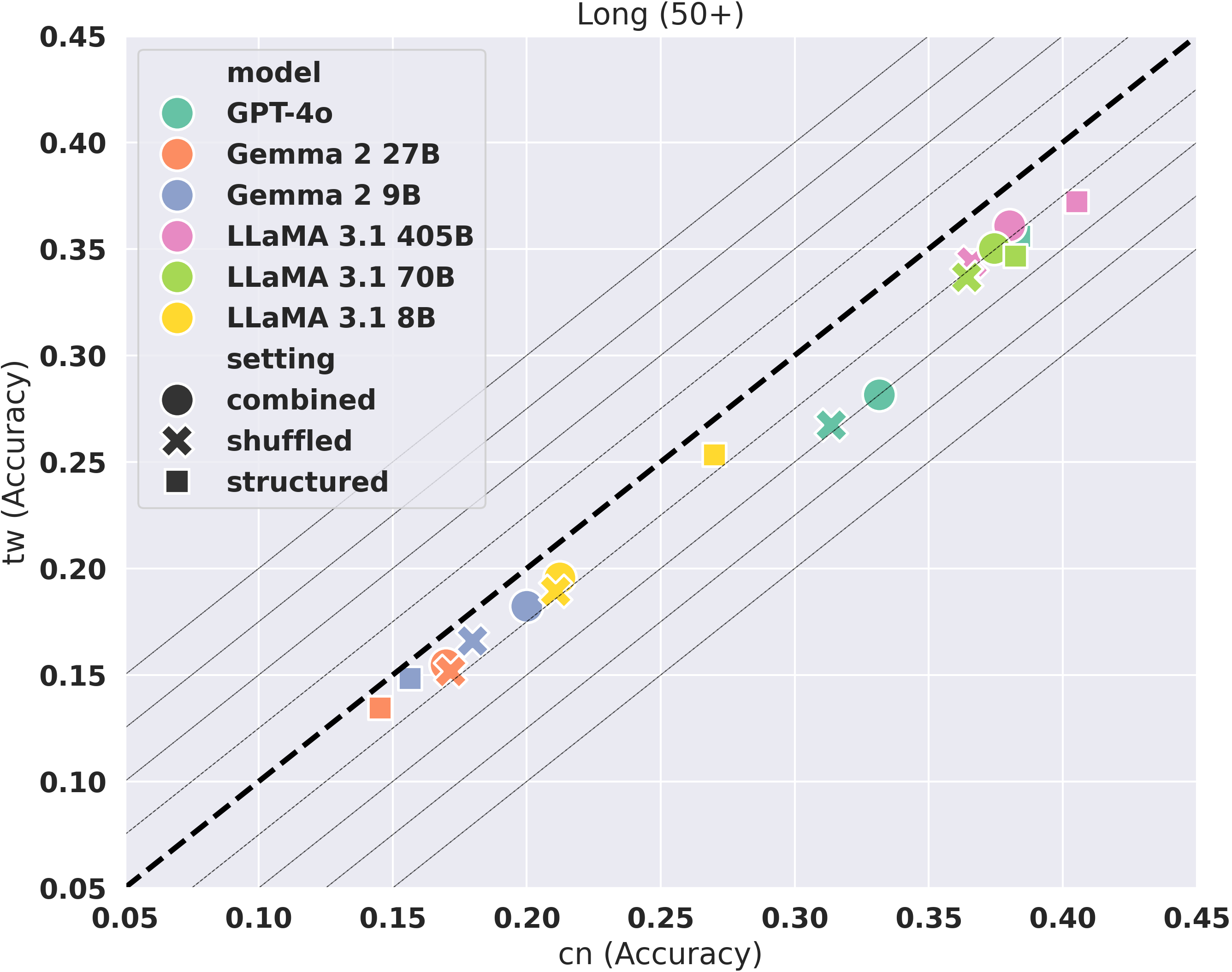}
    \caption{Comparison of accuracy between \cnChinese and \twChinese for short (left) and long (right) texts. Each point represents a [model, setting]'s performance. The diagonal line ($x=y$) indicates equal performance. Points above the line suggest better performance in \twChinese, while points below suggest better performance in \cnChinese. We do not see a big difference between the short and long texts.}
    \label{fig:main-lengh-analysis-accuracy}
\end{figure*}

\begin{figure*}
    \centering
    \includegraphics[width=0.48\linewidth]{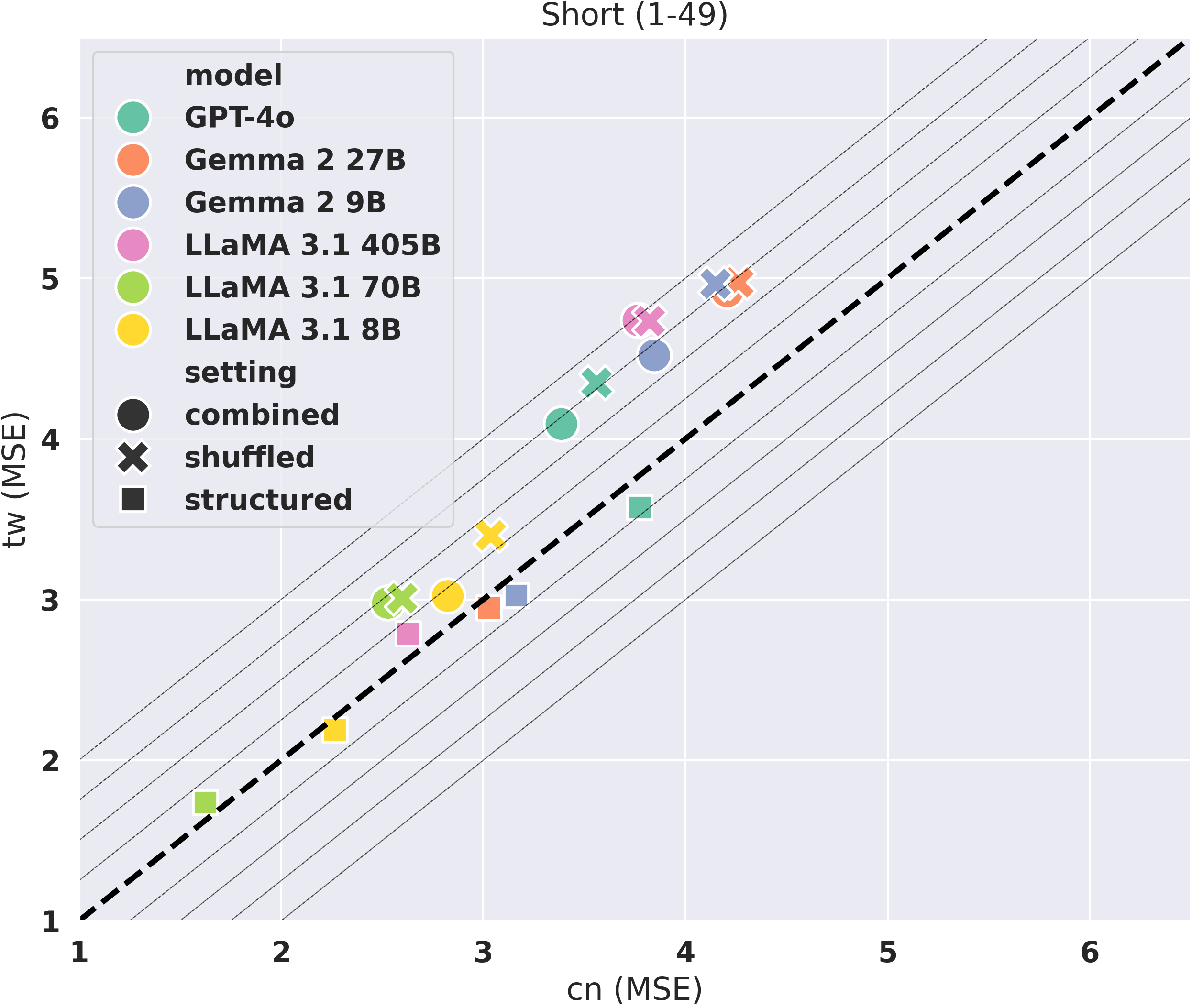}
    \hfill
    \includegraphics[width=0.48\linewidth]{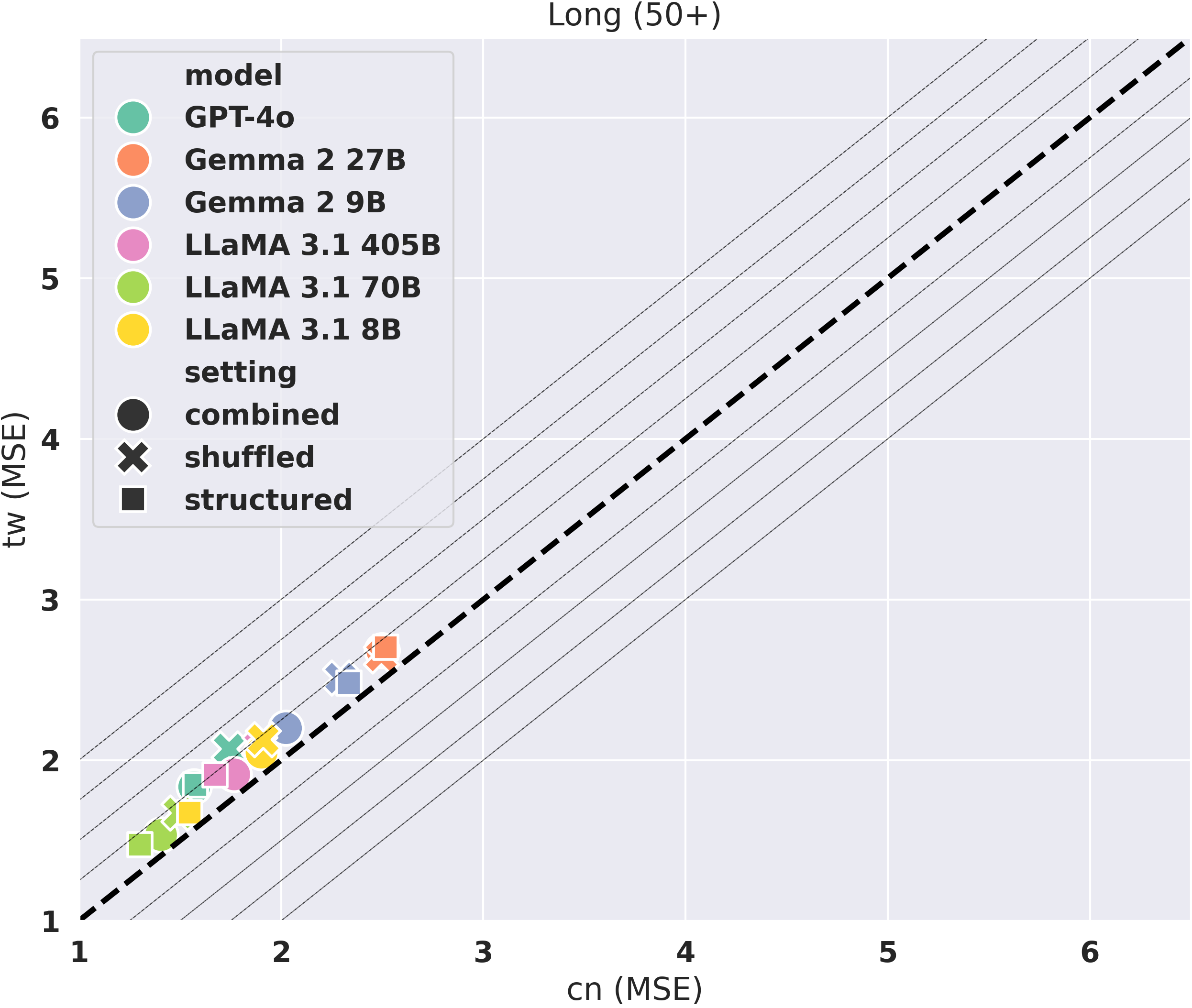}
    \caption{Comparison of MSE between \cnChinese and \twChinese for short (left) and long (right) texts. Each point represents a model's performance. The diagonal line ($x=y$) indicates equal performance. Points below the line suggest better performance in \twChinese, while points above suggest better performance in \cnChinese. Note the larger performance gap for short texts compared to long texts.}
    \label{fig:main-length-analysis-mse}
\end{figure*}

\section{Impact of Length on Model Performance\label{app:length-analysis-in-main-result}}
To further analyze the effect of text length on our main study results presented in \Cref{sec:experiment},
we plotted the performance on scatter plots.
The x-axis represents the performance for \cnChinese, while the y-axis represents the performance for \twChinese.
The results are displayed in \Cref{fig:main-lengh-analysis-accuracy} and \Cref{fig:main-length-analysis-mse}.

In these plots, the diagonal line ($x=y$) represents equal performance between the two language variations.
The distance of each point from this line indicates the performance gap.
For the accuracy plot (\Cref{fig:main-lengh-analysis-accuracy}), points closer to the bottom-right indicate better performance in \cnChinese, while points closer to the top-left indicate better performance in \twChinese.
Conversely, in the MSE plot (\Cref{fig:main-length-analysis-mse}), points closer to the top-left indicate better performance in \cnChinese.

Our analysis of \Cref{fig:main-lengh-analysis-accuracy} does not reveal a significant difference between the short and long text groups in terms of accuracy.
However, \Cref{fig:main-length-analysis-mse} shows a larger gap for the short text group compared to the long text group in terms of MSE. 
Based on these observations, we hypothesize that shorter reviews may introduce more bias. This could be due to insufficient contextual information in shorter texts, where models have to judge based on its prior knowledge.

\begin{table}[t]
  \includegraphics[width=\linewidth]{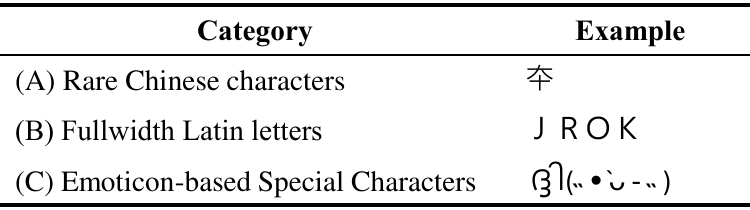}
  \caption{Example of special characters found in our dataset.}
  \label{tab:special-characters}
\end{table}

\begin{table}[]
\centering
\footnotesize
\addtolength{\tabcolsep}{-1.2mm}
\begin{tabular}{@{}l@{\kern0pt}rrrr@{}}
\toprule
\multirow{2}{*}{\textbf{Category}} & \multicolumn{2}{c}{\textbf{CN}} & \multicolumn{2}{c}{\textbf{TW}} \\ \cmidrule(lr){2-3} \cmidrule(lr){4-5}
 & \multicolumn{1}{c}{\textbf{Count}} & \multicolumn{1}{c}{\textbf{Ratio}} & \multicolumn{1}{c}{\textbf{Count}} & \multicolumn{1}{c}{\textbf{Ratio}} \\ \midrule
\textbf{Only Traditional} & 2,000 & 8.73\% & 17,130 & 74.74\% \\
\textbf{Only Simplified} & 15,816 & 69.01\% & 90 & 0.39\% \\
\textbf{Only English }& 107 & 0.47\% & 119 & 0.52\% \\
\textbf{Only Emoji }& 1 & 0.00\% & 4 & 0.02\% \\
\textbf{Only Symbol }& 1 & 0.00\% & 5 & 0.02\% \\
\textbf{Only Bopomofo} & 4 & 0.02\% & 35 & 0.15\% \\
\textbf{Only JP/KR} & 0 & 0.00\% & 0 & 0.00\% \\
\textbf{Only Punctuation} & 4 & 0.02\% & 5 & 0.02\% \\
\textbf{Only Unknown} & 0 & 0.00\% & 0 & 0.00\% \\ \midrule
\textbf{Traditional + English} & 251 & 1.10\% & 3,022 & 13.19\% \\
\textbf{Traditional + Emoji} & 75 & 0.33\% & 666 & 2.91\% \\
\textbf{Traditional + Symbol} & 79 & 0.34\% & 894 & 3.90\% \\
\textbf{Traditional + Bopomofo} & 8 & 0.03\% & 66 & 0.29\% \\
\textbf{Traditional + JP/KR} & 0 & 0.00\% & 9 & 0.04\% \\
\textbf{Traditional + Unknown} & 30 & 0.13\% & 246 & 1.07\% \\ \midrule
\textbf{Simplified + English} & 2,681 & 11.70\% & 12 & 0.05\% \\
\textbf{Simplified + Emoji} & 383 & 1.67\% & 1 & 0.00\% \\
\textbf{Simplified + Symbol} & 323 & 1.41\% & 0 & 0.00\% \\
\textbf{Simplified + Bopomofo} & 0 & 0.00\% & 0 & 0.00\% \\
\textbf{Simplified + JP/KR} & 22 & 0.10\% & 0 & 0.00\% \\
\textbf{Simplified + Unknown} & 90 & 0.39\% & 1 & 0.00\% \\ \bottomrule
\end{tabular}
\addtolength{\tabcolsep}{+1.2mm}
\caption{Language distribution. CN and TW users similarly mix non-Chinese elements with their primary writing systems (Simplified or Traditional Chinese). However, CN users incorporate Traditional characters more frequently than TW users use Simplified ones.}
\label{tab:language-distribution}
\end{table}

\section{Language Detection Analysis\label{appendix:language-analysis}}
To have a better understanding of Chinese and non-Chinese script elements in reviews, we conducted a detailed character-level analysis across our dataset. Using predefined vocabulary sets from zhon~\cite{githubzhon}, the Unicode Character Database~\cite{unicode}, and emoji~\cite{githubemoji}, we categorized characters into the following groups: traditional Chinese characters, simplified Chinese characters, English letters, emojis, bopomofo, Japanese characters, Korean characters, mathematical symbols, punctuation, and numbers. The table below presents the distribution of these elements across CN and TW users' reviews.

Our analysis revealed that CN and TW users exhibit similar patterns when incorporating non-Chinese elements into their primary writing system (Simplified Chinese with other elements for CN users, Traditional Chinese with other elements for TW users). The key difference lies in cross-script usage: CN users demonstrate a higher frequency of Traditional character usage compared to TW users' usage of Simplified characters.

Beyond the identified script elements, we found 103 characters in an ``Unknown'' category, appearing across 388 samples. Further investigation revealed these primarily consist of (1) rare Chinese characters not included in the zhon~\cite{githubzhon} vocabulary list (\ref{tab:special-characters} (A)), (2) fullwidth Latin letters (\ref{tab:special-characters} (B)), and (3) characters from other languages, with the latter mainly used in emoticons (\ref{tab:special-characters} (C)). As our current analysis is conducted at the character level, we cannot identify complete pinyin words or emoticon compositions. We will acknowledge this limitation and encourage future research to explore these aspects more comprehensively.

\paragraph{How Non-Chinese Elements Affect LLM Performance?}
To investigate how non-Chinese elements affect LLM performance,
we analyzed GPT-4o's performance on review pairs under different language constraints.
We define ``Chinese'' as the primary writing system for each user group
(Traditional for \twChinese users, Simplified for \cnChinese users).
We included only pairs where both reviews strictly adhered to these constraints.
For instance, \cnChinese reviews must contain only Simplified Chinese characters,
while \twChinese reviews must contain only Traditional Chinese characters.
``Chinese+English'' refers to reviews containing only the primary Chinese writing system plus English letters.

The results are presented in \ref{tab:language-subset-exp}.
When restricting the analysis to primary Chinese characters only (the Chinese row),
the performance gap between \twChinese and \cnChinese widened (see [plain, $\Delta$MSE] and [shuffled, $\Delta$MSE]),
indicating a potential bias in processing Traditional versus Simplified Chinese characters.
In the code-switching scenario with English letters, both groups showed relatively closer performance, with a smaller gap between them. This suggests that English elements may help normalize the performance across both language groups.


\begin{table*}[]
\footnotesize
\centering
\begin{tabular}{cccccr@{\kern-11pt}lccr@{\kern-11pt}l}
\toprule
 &  &  & \multicolumn{4}{c}{\textbf{Acc$\uparrow$}} & \multicolumn{4}{c}{\textbf{MSE$\downarrow$}} \\ \cmidrule(lr){4-7} \cmidrule(lr){8-11} 
\multirow{-2}{*}{\textbf{Setting}} & \multirow{-2}{*}{\textbf{Char. Set}} & \multirow{-2}{*}{\textbf{\#Pairs}} & \textbf{tw} & \textbf{cn} & \multicolumn{2}{c}{\textbf{\begin{tabular}[c]{@{}c@{}}$\Delta$Acc\\ (cn-tw)\end{tabular}}} & \textbf{tw} & \textbf{cn} & \multicolumn{2}{c}{\textbf{\begin{tabular}[c]{@{}c@{}}$\Delta$MSE\\ (cn-tw)\end{tabular}}} \\ \midrule
\textbf{structured} & \textbf{All} & 22,918 & 29.614 & 31.172 & \cellcolor[HTML]{F8D7D4}1.558\hspace{4mm} & *** & 2.985 & 3.026 & \cellcolor[HTML]{DBF3F4}0.206\hspace{4mm} &   \\
\textbf{structured} & \textbf{Chinese} & 12,237 & 28.193 & 29.901 & \cellcolor[HTML]{F7D3D0}1.708\hspace{4mm} & ** & 2.965 & 3.013 & \cellcolor[HTML]{F1FAFB}0.082\hspace{4mm} &   \\
\textbf{structured} & \textbf{Chinese+English} & 917 & 37.514 & 37.077 & \cellcolor[HTML]{EEF9FA}-0.436\hspace{4mm} & \multicolumn{1}{l}{} & 1.762 & 1.700 & \cellcolor[HTML]{FCF1F0}-0.107\hspace{4mm} &   \\ \midrule
\textbf{plain} & \textbf{All} & 22,914 & 22.231 & 25.011 & \cellcolor[HTML]{F2B7B2}2.780\hspace{4mm} & *** & 3.323 & 2.768 & \cellcolor[HTML]{FBECEB}-0.147\hspace{4mm} & *** \\
\textbf{plain} & \textbf{Chinese} & 12,237 & 21.051 & 24.197 & \cellcolor[HTML]{F0ADA8}3.146\hspace{4mm} & *** & 3.335 & 2.642 & \cellcolor[HTML]{E7F7F8}0.138\hspace{4mm} & *** \\
\textbf{plain} & \textbf{Chinese+English} & 917 & 28.571 & 30.862 & \cellcolor[HTML]{F4C4C0}2.290\hspace{4mm} & \multicolumn{1}{l}{} & 1.943 & 1.799 & \cellcolor[HTML]{F1FAFB}0.083\hspace{4mm} &   \\ \midrule
\textbf{shuffled} & \textbf{All} & 22,915 & 21.353 & 24.002 & \cellcolor[HTML]{F2BAB5}2.649\hspace{4mm} & *** & 3.573 & 2.941 & \cellcolor[HTML]{F8DDDB}-0.269\hspace{4mm} & *** \\
\textbf{shuffled} & \textbf{Chinese} & 12,237 & 20.315 & 22.857 & \cellcolor[HTML]{F3BDB8}2.542\hspace{4mm} & *** & 3.580 & 2.808 & \cellcolor[HTML]{EC9F98}-0.772\hspace{4mm} & *** \\
\textbf{shuffled} & \textbf{Chinese+English} & 917 & 26.609 & 28.680 & \cellcolor[HTML]{F5C9C6}2.072\hspace{4mm} & \multicolumn{1}{l}{} & 2.196 & 1.937 & \cellcolor[HTML]{F8DEDC}-0.260\hspace{4mm} &  \\ \bottomrule
\end{tabular}
\caption{Analysis of LLM performance across different character sets.}
\label{tab:language-subset-exp}
\end{table*}

\end{document}